# Supervised Texture Segmentation: A Comparative Study

Omar S. Al-Kadi
King Abdullah II School for IT
University of Jordan, Amman, 11945 JORDAN
o.alkadi@ju.edu.jo

*Abstract-* **This paper aims to compare between four different types of feature extraction approaches in terms of texture segmentation. The feature extraction methods that were used for segmentation are Gabor filters (GF), Gaussian Markov random fields (GMRF), run-length matrix (RLM) and co-occurrence matrix (GLCM). It was shown that the GF performed best in terms of quality of segmentation while the GLCM localises the texture boundaries better as compared to the other methods.**

## I. INTRODUCTION

One of the challenging aspects of machine vision is being able to robustly distinguish between different objects. The more objects appear in the scene the harder the task becomes. Therefore researchers in computer vision have developed many methods to detect variation in intensities ─ that might be, but not necessarily, due to different objects ─ through extracting texture features (i.e. texture measures), then using a classification approach to automate the process.

This work compares the quality of segmentation of four different types of texture measures which have been used for image analysis and segmentation [1, 2]. We used two statistical based methods represented by co-occurrence and run-length matrices, and a model and signal processing method represented by Markov random fields and Gabor filters; respectively. Then the images were segmented using a naïve Bayesian classifier and the quality of segmentation was measured using the Bhattacharyya distance.

## II. FEATURE EXTRACTION

### A. Markov random fields

First, based upon the Markovian property, which is simply the dependence of each pixel in the image on its neighbours only, and using a Gaussian Markov random field model (GMRF) for third order Markov neighbours [3], the GMRF parameters are estimated using least square error estimation method. The GMRF model is defined by the following formula:

$$p(I_{ij}|I_{kl},(k,l) \in N_{ij}) = \frac{1}{\sqrt{2\pi\sigma^2}} \exp\left\{\frac{I_{ij} - \sum_{l=1}^{n}\alpha_l s_{kl;l}}{2\sigma^2}\right\} \quad (1)$$

where the right hand side of equation (1) represents the probability of a pixel $(i, j)$ having a specific grey value $I_{ij}$, given the values of its neighbours, $n$ is the total number of pixels in the neighbourhood $N_{kl}$ of pixel $I_{ij}$, which influence its value, $\alpha_l$ is the parameter with which a neighbour influences the value of $(i, j)$, and $s_{kl;l}$ is the value of the pixel at the corresponding position, where

$$s_{ij;1} = I_{i-1,j} + I_{i+1;j}$$
$$s_{ij;2} = I_{i,j-1} + I_{i;j+1}$$
$$s_{ij;3} = I_{i-2,j} + I_{i+2;j}$$
$$s_{ij;4} = I_{i,j-2} + I_{i;j+2}$$
$$s_{ij;5} = I_{i-1,j-1} + I_{i+1;j+1}$$
$$s_{ij;6} = I_{i-1,j+1} + I_{i+1;j-1} \quad (2)$$

The MRF parameters $\alpha$ and $\sigma$ are estimated using least square error estimation method, as follows:

$$\begin{pmatrix}\alpha_l \\ \vdots \\ \alpha_n\end{pmatrix} = \left\{\sum_{ij}\begin{bmatrix}s_{ij;1}s_{ij;1} & \cdots & s_{ij;1}s_{ij;n} \\ \vdots & \ddots & \vdots \\ s_{ij;n}s_{ij;1} & \cdots & s_{ij;n}s_{ij;n}\end{bmatrix}\right\}^{-1} \sum_{ij}I_{ij}\begin{pmatrix}s_{ij;1} \\ \vdots \\ s_{ij;n}\end{pmatrix} \quad (3)$$

$$\sigma^2 = \frac{\sum_{ij}\left[I_{ij} - \sum_{l=1}^{n}\alpha_l s_{ij;l}\right]}{(M-2)(N-2)} \quad (4)$$

where $M$ and $N$ are the size of the image.

## B. Gabor filters

The Gabor filter (GF) is a Gaussian modulated sinusoidal with a capability of multi-resolution decomposition due to its localization in the spatial and spatial-frequency domain. Jain and Farrokhnia used a dyadic Gabor filter bank covering the spatial-frequency domain with multiple orientations [2]. The real impulse response of a 2-D sinusoidal plane wave with orientation $\theta$ and radial centre frequency $f_0$ modulated by a Gaussian envelope with standard deviations $\sigma_x$ and $\sigma_y$ is given by

$$h(x,y) = \exp\left\{-\frac{1}{2}\left[\frac{x^2}{\sigma_x^2} + \frac{y^2}{\sigma_y^2}\right]\right\} \cdot \cos(2\pi f_0 x) \quad (5)$$

where  $x' = x\cos\theta + y\sin\theta$
 $y' = -x\sin\theta + y\cos\theta$

Figure 1 shows the frequency response of the dyadic filter bank in the spatial-frequency domain. For the images having size of 256 x 256 used in this work, five radial frequencies ($\sqrt{2}/2^6, \sqrt{2}/2^5, \sqrt{2}/2^4, \sqrt{2}/2^3, and \sqrt{2}/2^2$) with 4 orientations ($0^0, 45^0, 90^0, 135^0$) was adopted according to [2]. Finally the extracted features would represent the energy of each magnitude response.

## C. Co-occurrence matrix

The Grey level co-occurrence matrix (GLCM) represents the joint probability of certain sets of pixels having certain grey-level values. It calculates how many times a pixel with grey-level $i$ occurs jointly with another pixel having a grey value $j$. We can generate as much as $M$ x $N$ - 1 GLCMs with different directions by varying the displacement vector $d$ between each pair of pixels. For each image and with distance set to one, four GLCMs having directions ($0^0, 45^0, 90^0, 135^0$) were generated. Having the GLCM normalised, we can then derived eight second order statistic features which are also known as Haralick features [4] for each sample, which are: contrast, correlation, energy, entropy, homogeneity, dissimilarity, inverse difference momentum, maximum probability.

## D. Run-length matrix

The grey level run-length matrix (RLM) $P_r(i,j|\theta)$ is defined as the numbers of runs with pixels of gray level $i$ and run length $j$ for a given direction $\theta$ [5]. RLMs was generated for each image having directions ($0^0, 45^0, 90^0, 135^0$), then the following five statistical features were derived: short run emphasis, long run emphasis, gray level non-uniformity, run length non-uniformity and run percentage.

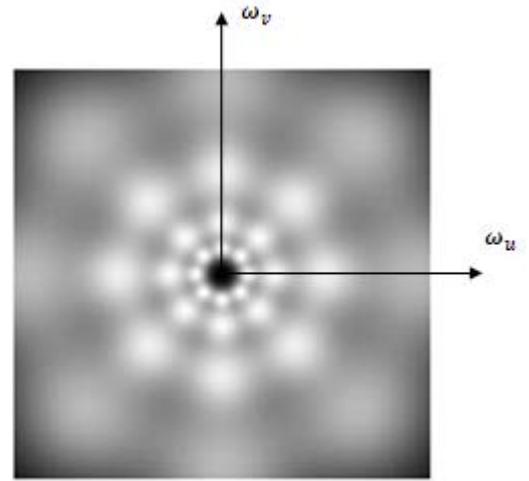

Figure 1. Gabor filter defined in the spatial-frequency domain with 45° orientation separation.

### III. PATTERN SEGMENTATION

Supervised segmentation was applied using a naïve Bayesian classifier (nBc) which is a simple probabilistic classifier that assumes attributes are independent. Yet, it is a robust method with on average has a good classification accuracy performance, and even with possible presence of dependent attributes [6].

From Bayes' theorem,

$$P_i(C_i/X) = \frac{P(X/C_i)P(C_i)}{P(X)} \quad (6)$$

Given a data sample $X$ which represent the extracted texture features vector $(f_1, f_2, f_3 \ldots f_j)$ having a probability density function (PDF) $P(X/C_i)$, we tend to maximize the posterior probability $P(C_i/X)$ (i.e., assign sample $X$ to the class $C_i$ that yields the highest probability value).

Where $P(C_i/X)$ is the probability of assigning class $i$ given feature vector $X$; and $P(X/C_i)$ is the probability; $P(C_i)$ is the probability that class $i$ occurs in all data set; $P(X)$ is the probability of occurrence of feature vector $X$ in the data set.

$P(C_i)$ and $P(X)$ can be ignored since we assume that all are equally probable for all samples. Which yields that the maximum of $P(C_i/X)$ is equal to the maximum of $P(X/C_i)$ and can be estimated using maximum likelihood after assuming a Gaussian PDF [7] as follows:

$$P(X/C_i) = \frac{1}{(2\pi)^{n/2}|\Sigma_i|^{1/2}} exp\left[-\frac{1}{2}(X-\mu_i))^T \sum_i^{-1} (X-\mu_i)\right] \quad (7)$$

where $\Sigma_i$ and $\mu_i$ are the covariance matrix and mean vector of feature vector $X$ of class $C_i$; $|\Sigma_i|$ and $\Sigma_i^{-1}$ are the determinant and inverse of the covariance matrix; and $(X - \mu_i)^T$ is the transpose of $(X - \mu_i)$.

## IV. MEASURING SEGMENTATION QUALITY

The Bhattacharyya distance was used to assess the quality of the segmentation by measuring the difference between each segmented texture and its reference image. This method calculates the upper bound of classification error between feature class pairs [8], indicating the smaller the distance the better the better the quality of segmentation.

$$B_{c_1 c_2} = \frac{1}{8}(\mu_1 - \mu_2)^T \left(\frac{\Sigma_1 + \Sigma_2}{2}\right)^{-1} (\mu_1 - \mu_2) + \frac{1}{2} ln\left(\frac{|\Sigma_1 + \Sigma_2|}{2\sqrt{|\Sigma_1||\Sigma_2|}}\right) \quad (8)$$

where $|\Sigma_i|$ is the determinant of $\Sigma_i$, and $\mu_i$ and $\Sigma_i$ are the mean vector and covariance matrix of class $C_i$

## V. EXPERIMENTAL RESULTS AND DISCUSSION

The four texture measures were applied to Nat-5 image which is composed of five different types of textures from the Brodatz album [9]. Using supervised segmentation; each pixel with a certain feature vector in the Nat-5 image was assigned a class label by applying the trained nBc. Observing the segmented images, the GLCM and the GF gave a better performance in capturing the characteristics of the different image textures; while RLM method showed that it is not as effective, with many misclassification errors in the top and bottom textures (see Figure 2).

The Bhattacharyya distance was used to quantitatively assess the quality of segmentation for each of the five different textures in the Nat-5 image. Each of the segmented regions was measured against its corresponding pair in the original image, and then all distances were summed up to give an assessment of the quality of segmentation. Table I shows the GF having the best segmentation (the least difference between the segmented and reference texture) as compared to other methods, although the GLCM tends to localise the texture boundaries better as compared to the other methods (see Figure 3), it came third in the segmentation assessment due to the misclassification of *texture1* as compared to the classification accuracy of the remaining four textures. Nonetheless, noise that might affect the quality of the extracted features needs to be investigated.

In order to confirm the results, experiments were also performed on 78 different Brodatz textures as shown in Table II, which were categorised into texture-pairs (S-ioa, S-iob, and S-ioc), five-texture synthesis (Nat-5a, Nat-5b, Nat-5c, and Nat-5d), ten-texture synthesis (Nat-10a and Nat-10b), and sixteen-texture synthesis (Nat-16a and Nat-16b).

It is obvious that the quality of the segmented images decrease as the number of textures in the image increase. This might be due to the fixed size 32 x 32 pixels sliding window used in the segmentation process for each of the 256 x 256 pixel texture images used in this paper. That is, the more texture synthesised in the fixed size image, the smaller the area each texture occupies (i.e. the sliding window could simultaneously cover multiple textures, especially at texture boarders), and thus the harder the task of the feature extraction method becomes. Future work would focus on studying the effect of the sliding window size on segmentation quality, and the robustness of the feature extraction methods when applied to complex out-door scenes, which is considered a more challenging problem.

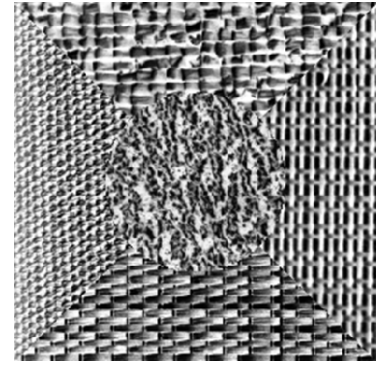
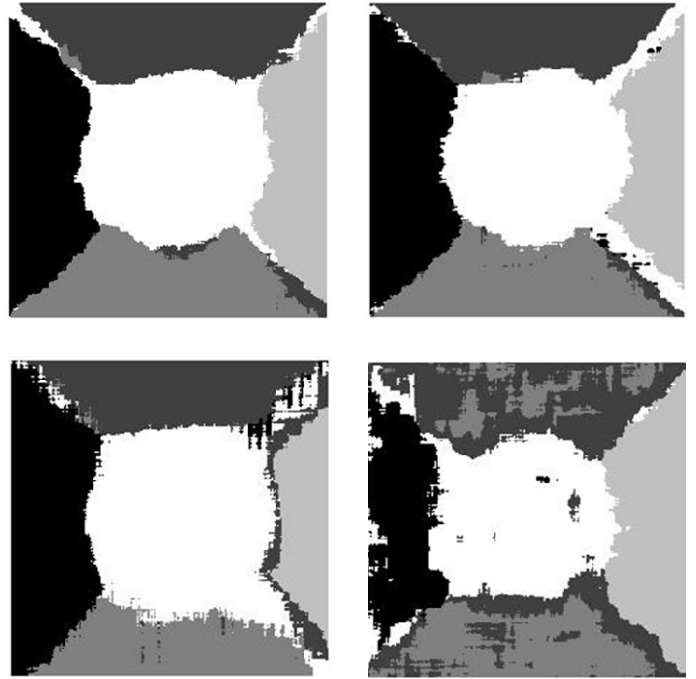

Figure 2. At the top is the Nat-5 image to be segmented, while the four images beneath it, and from left to right and from top to bottom are the segmented images using GLCM, GF, GMRF and RLM; respectively.

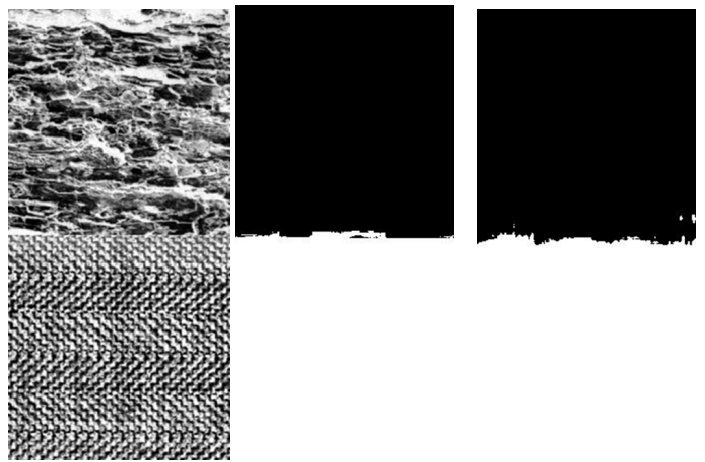

Figure 3. From left to right, a two-texture image followed by segmentation results (black for the upper and white for the lower texture) using GLCM and GF, respectively.

Table I. Bhattacharyya distance for each of the five
textures in the Nat-5 image referring to four texture measure

| Texture measure | Texture image | | | | | Total distance |
|---|---|---|---|---|---|---|
| | *texture 1* | *texture 2* | *texture 3* | *texture 4* | *texture 5* | |
| GLCM | $1.27 \times 10^{-1}$ | $2.50 \times 10^{-2}$ | $7.28 \times 10^{-4}$ | $7.40 \times 10^{-4}$ | 0 | $1.52 \times 10^{-1}$ |
| GF | $6.80 \times 10^{-2}$ | $2.20 \times 10^{-2}$ | $1.00 \times 10^{-3}$ | $2.00 \times 10^{-3}$ | 0 | $9.30 \times 10^{-2}$ |
| GMRF | $1.30 \times 10^{-1}$ | $1.10 \times 10^{-2}$ | $1.00 \times 10^{-3}$ | $2.98 \times 10^{-4}$ | $1.78 \times 10^{-9}$ | $1.42 \times 10^{-1}$ |
| RLM | $2.49 \times 10^{-1}$ | $2.00 \times 10^{-2}$ | $1.00 \times 10^{-3}$ | $7.92 \times 10^{-6}$ | $3.16 \times 10^{-5}$ | $2.70 \times 10^{-1}$ |

Table II. Original versus segmented results for the GLCM, MRF, RLM, and GF feature extraction methods using 78 different Brodatz textures synthesised in paired-textures (first three rows), five-textures (row four to row seven), ten-textures (row 8 and 9), and sixteen-textures (last two rows).

| Original image | | Segmented image | | | |
|---|---|---|---|---|---|
| Texture kind | Texture synthesis | Feature extraction method | | | |
| | | GLCM | MRF | RLM | GF |
| S-ioa | 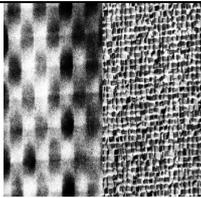 | 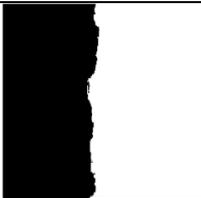 | 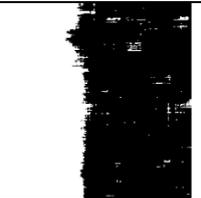 | 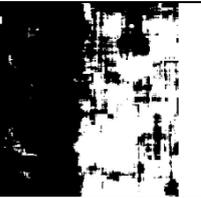 | 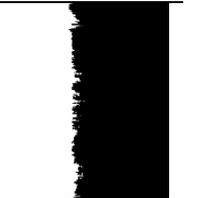 |
| S-iob | 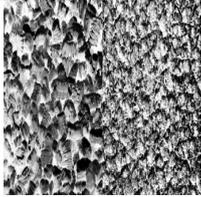 | 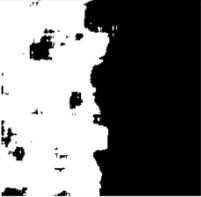 | 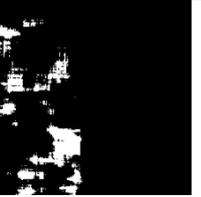 | 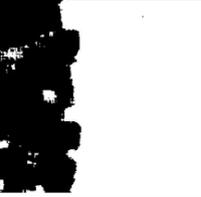 | 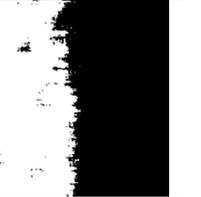 |
| S-ioc | 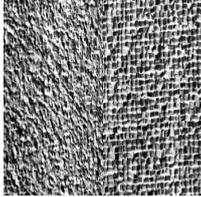 | 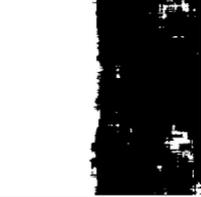 | 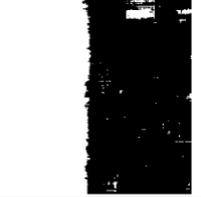 | 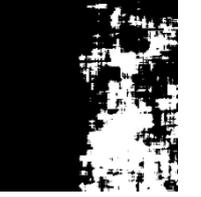 | 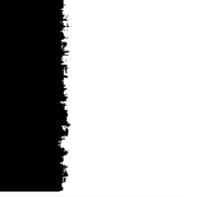 |
| Nat-5a | 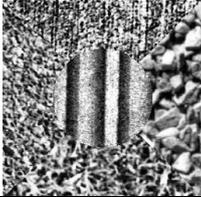 | 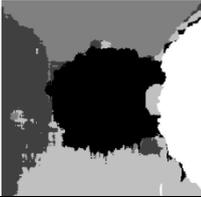 | 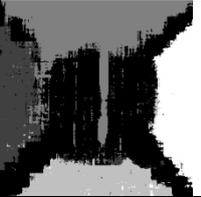 | 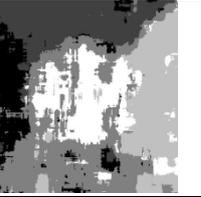 | 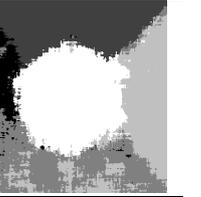 |
| Nat-5b | 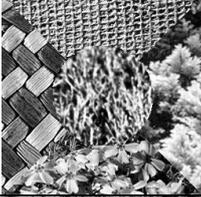 | 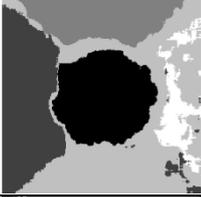 | 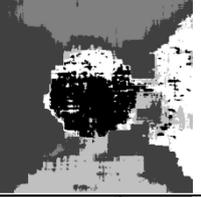 | 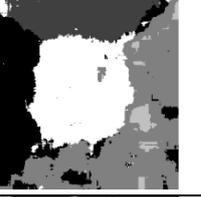 | 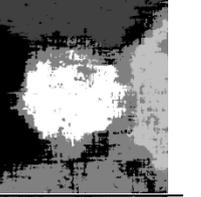 |
| Nat-5c | 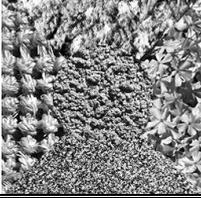 | 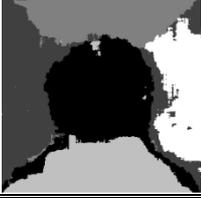 | 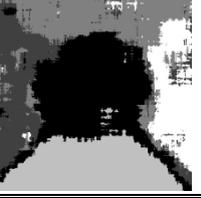 | 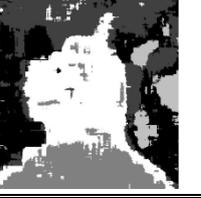 | 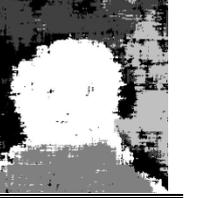 |

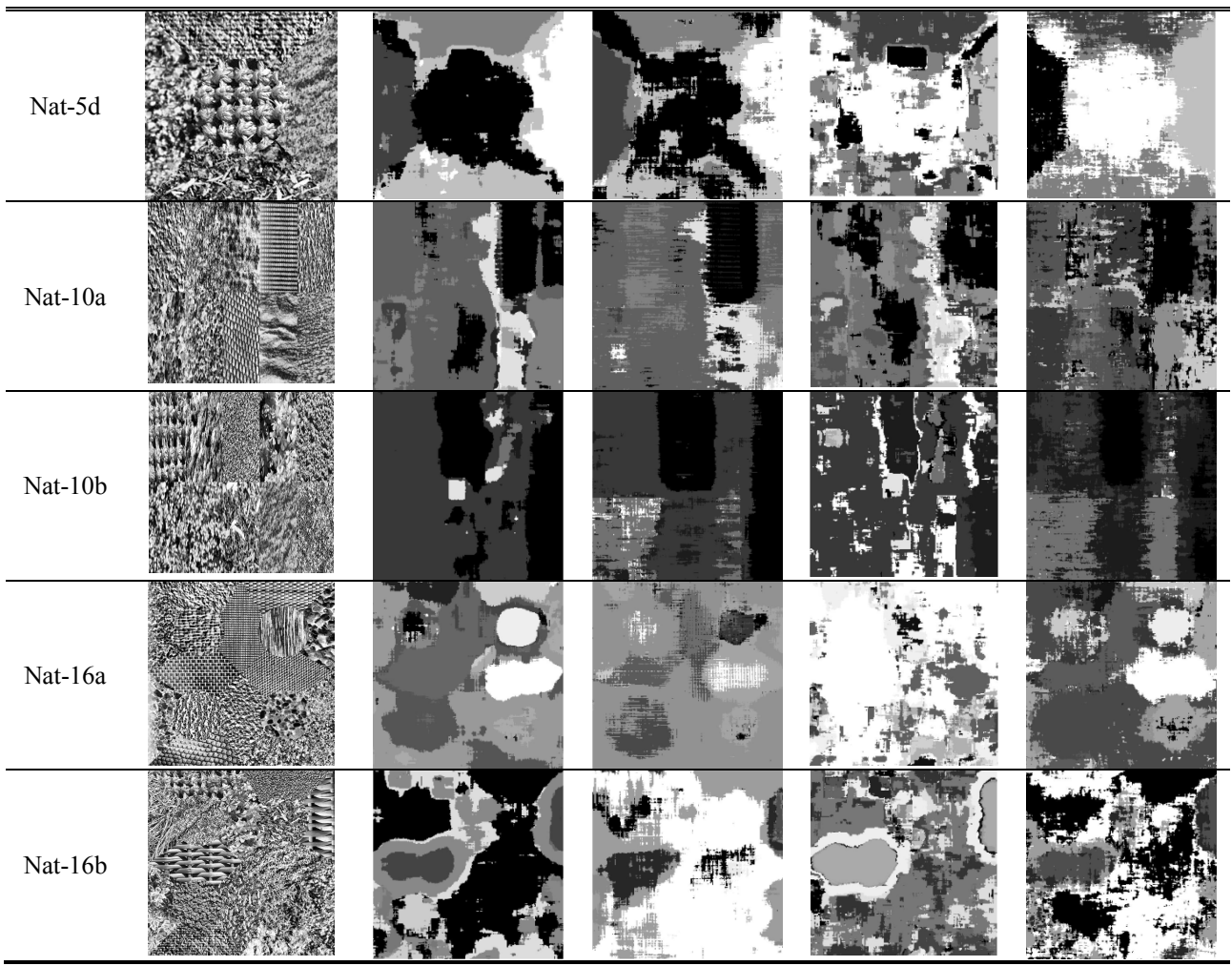

## VI. CONCLUSION

As various texture measures capture different characteristics of the same texture, the segmentation performance varies from one method to another depending on the robustness of these features. The quality of the segmentation was compared using four different texture measures applied to a five texture image. The GF gave less segmentation error, yet the GLCM was better in localising texture boundaries, while the RLM showed that it does not characterise texture efficiently as the rest.